# SAVIOR: Sample-efficient Adaptation of Vision-Language Models for OCR Representation

Akshata A Bhat[1,*], Sharath Naganna[1,2], Saiful Haq[1,2], Prashant Khatri[1],
Krishna Chaitanya Reddy Tamataam[1], Neha Arun[1], Niyati Chhaya[1], Pushpak Bhattacharyya[2]

[1]Hyperbots Inc
[2]Indian Institute of Technology Bombay

## Abstract

*Optical Character Recognition (OCR) pipelines and vision–language models exhibit high accuracy on standard OCR and document-understanding benchmarks, yet they systematically underperform on document patterns that are operationally critical in financial workflows. These patterns include vertical or rotated text, logo-embedded vendor names, fine-print regulatory clauses, degraded scans, and complex multi-column layouts; while underrepresented in public datasets, they constitute a substantial portion of real-world failure cases. We introduce SAVIOR, a sample-efficient data curation methodology that targets such high-impact failure scenarios to adapt vision–language models for robust financial OCR. We further propose PaIRS, a structure-aware evaluation metric that measures layout fidelity by comparing pairwise spatial relationships between tokens. Fine-tuning Qwen2.5-VL-Instruct with SAVIOR-Train achieves a word recall of 0.93, and when the resulting layout-aware OCR output is used as context for document question answering, it attains an F1 score of 0.89, outperforming both open and closed-source models such as GPT-4o, Mistral-OCR, PaddleOCR-VL and DeepSeek-OCR.*

## 1. Introduction

Modern document-processing pipelines rely heavily on OCR, using it either as a standalone text-extraction component or as part of larger information-retrieval and question-answering workflows. As these applications have expanded to include increasingly varied document types, traditional modular OCR pipelines have been gradually replaced by standalone transformer-based vision–language models, including both closed-source systems (e.g., GPT-4o [22], Mistral OCR [1], Gemini-2.5-Flash) and open-source models (e.g., PaddleOCR 3.0 [4], PaddleOCR-VL [3], Nanonets-OCR-s [20]), which generate text directly from images and jointly learn visual and textual representations. These models achieve strong performance on public benchmarks and general-purpose documents, yet they often struggle with domain-specific visual patterns that occur only rarely in their training data. As a result, they typically capture most of the content but frequently miss or misinterpret the specialized visual tokens characteristic of many real-world documents.

These challenges become especially visible in financial documents, where domain-specific visual patterns are both common and critical. Invoices, receipts, statements, and compliance forms contain visual elements that are essential for accurate interpretation but appear infrequently in the datasets used to train most OCR and vision–language models. These include vertical or rotated field labels, logo-embedded vendor names, dense multi-column layouts, fine-print regulatory clauses, handwritten annotations, and degraded scans produced by legacy systems. Figure 1 shows a document which contains different failure cases. Errors in these elements can break reading order, disrupt key–value extraction, and weaken downstream reasoning in accounting or auditing workflows, even when most of the surrounding content is transcribed correctly. As a result, reliable OCR performance in financial settings requires robustness to a set of domain-specific visual tokens that standard models consistently fail to handle.

A key reason for these persistent errors lies in how existing OCR and vision–language models represent document structure. Many systems rely on layout-aware outputs that inject bounding boxes, markup tags, or special formatting tokens to encode spatial arrangement. Figure 2 illustrates the failure of Nanonets-OCR-s on a synthetic sample representative of a typical financial document. While this captures structure explicitly, it often disrupts natural reading order and requires additional post-processing to recover a human-readable sequence. Other models favor plain reading-order output, which preserves human interpretability but may lose fine-grained spatial relationships that are important for downstream extraction. Neither approach alone is sufficient for financial documents, where accurate interpretation depends on both correctly transcribed text and the preservation of meaningful spatial cues.

*Corresponding author
[1]https://mistral.ai/news/mistral-ocr

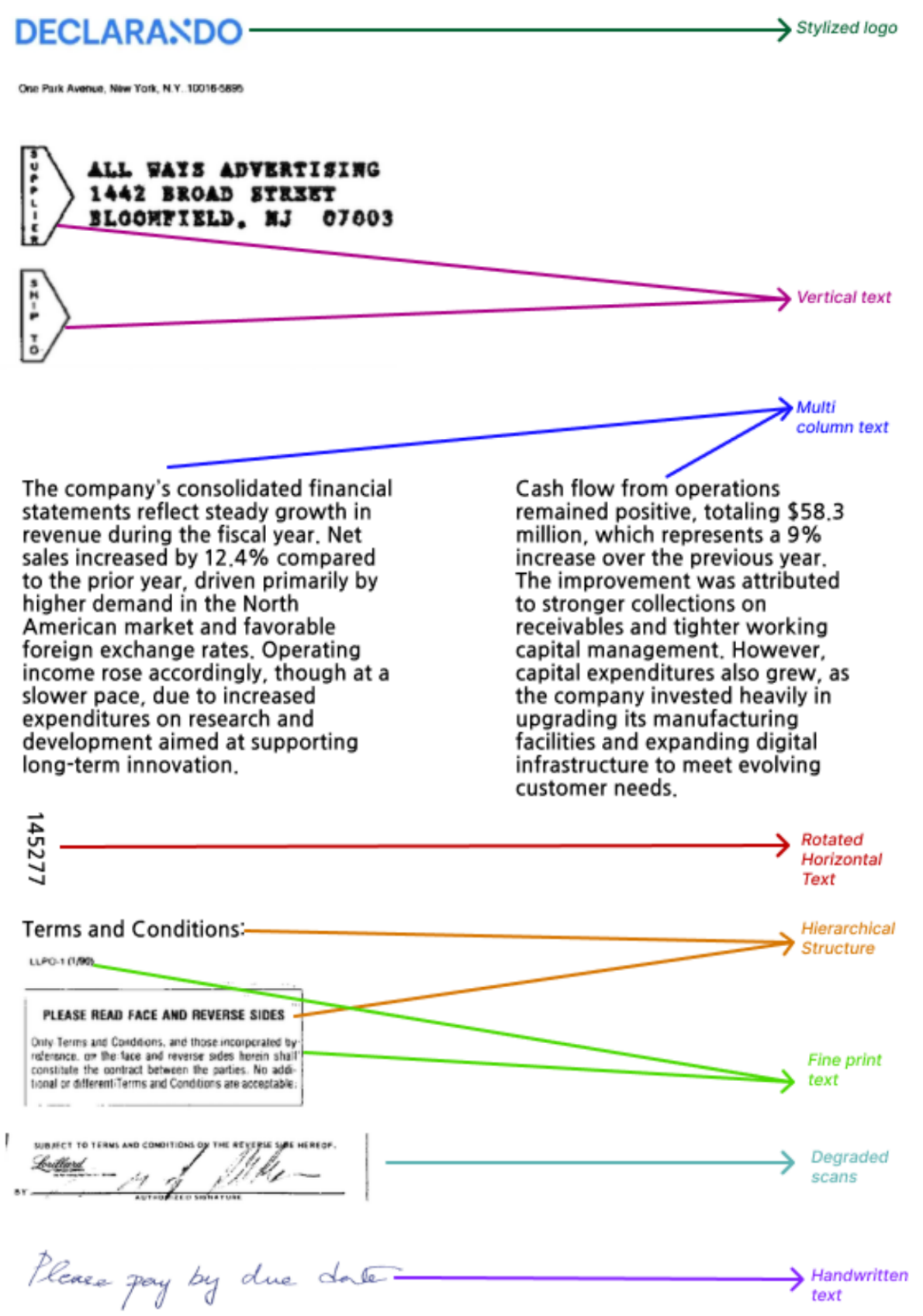

Figure 1. Example document illustrating several failure scenarios targeted in SAVIOR-Train, including vertical text, logo embedded fields, fine print, multi-column layout, and handwritten annotations.

To address these challenges, we introduce SAVIOR, a sample-efficient data curation strategy for adapting pretrained vision language models to the visual and structural patterns that frequently disrupt OCR in financial documents. SAVIOR focuses on the targeted selection of documents that contain recurring failure-inducing elements such as vertical text, logo-embedded fields, fine print clauses, noisy scans, and complex layout structures. Instead of expanding training sets indiscriminately, SAVIOR increases the concentration of difficult visual patterns inside a compact dataset so that adaptation is driven by the cases where pretrained models typically fail rather than by dataset scale.

Our contributions are as follows:

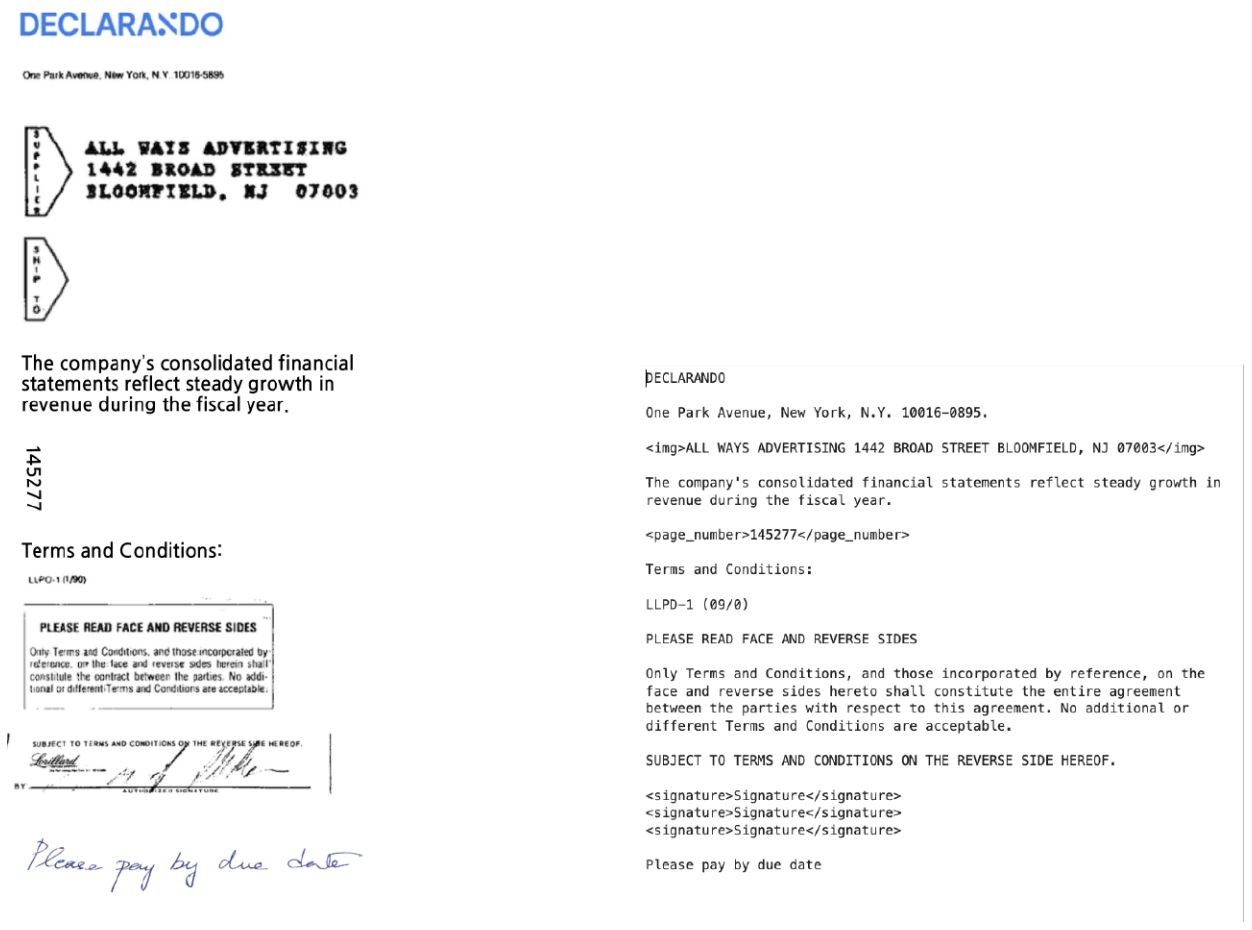

(a) Original document (b) OCR output

Figure 2. Comparison of the original document (left) with the OCR output (right). The OCR result fails to capture several critical elements such as vertical text, logo-embedded entities and fine-print clauses.

- SAVIOR: a sample-efficient approach for adapting vision–language models to OCR in financial documents, yielding robustness without large-scale corpus expansion.
- SAVIOR-Bench (509 expert-annotated documents), designed to evaluate transcription robustness and structural fidelity in financial OCR.[2]
- PaIRS, a layout-aware evaluation metric that quantifies structural fidelity by measuring pairwise relational similarity between predicted and ground-truth word positions, capturing both spatial correctness and downstream usability.

## 2. Related Work

Document understanding has advanced substantially with the development of multimodal and layout aware models. Early OCR based pipelines, including Tesseract [26], EasyOCR, and PP-OCR [5], follow a modular procedure of text detection, text recognition, and heuristic layout analysis. Although efficient and interpretable, these systems tend to degrade under real world conditions, such as skew, low contrast, occlusions, multilingual footers, and dense stamps [13, 25]. They also have limited contextual reasoning ability, which restricts performance on tasks like key value extraction and document question answering.

Multimodal vision language models (VLMs) were introduced to address these limitations. LayoutLM [29] and LayoutLMv3 [11] encode spatial structure using two-dimensional positional embeddings and achieve strong results on form-centric datasets such as FUNSD [12] and

[2] Only the test-set document images from SAVIOR-Bench are available on request (niyati.chhaya@hyprbots.com)

CORD [23], but they remain dependent on external OCR. Encoder-decoder architectures such as Donut [14] remove the OCR component entirely and generate structured outputs directly from pixels. Later work, including UDOP [27] and DocOwl [9], improves visual and textual alignment and supports instruction following for document question answering. More general purpose VLMs such as BLIP 2 [17], LLaVA [18], and InternVL [2] show competitive performance on OCRBench v2 [19] and DocVQA [21] without domain specific training.

In parallel, pipelines that combine OCR with large language models have demonstrated advantages on tasks requiring compositional or numerical reasoning. Recent studies [10, 28] have shown that high-quality OCR, together with models such as GPT-4o, Claude 3, or Qwen2.5, can outperform VLMs on enterprise workloads involving table fidelity, cross-document reference resolution, and metadata interpretation. These pipelines also benefit from modularity and ease of debugging, which are important considerations in regulated deployment settings.

Despite this progress, both VLM based approaches and OCR plus LLM approaches continue to struggle with enterprise specific visual anomalies. Public benchmarks rarely include vertical or low contrast text, nonstandard fonts, logo interference, legal microtext, or stamps embedded within tables. Datasets such as SynthDoG [14] and IIT CDIP [16] are valuable resources for document understanding research, but they do not reflect the visual noise, layout irregularities, and semantic variability that occur in modern enterprise invoices. Model efficiency research, including quantization [6, 30] and parameter efficient finetuning [1, 8], has improved deployability, yet evaluations still focus primarily on standard layouts.

# 3. SAVIOR-Train and SAVIOR-Bench

SAVIOR-TRAIN and SAVIOR-Bench were developed to improve OCR robustness in financial documents by concentrating on layout and transcription patterns that pretrained vision language models consistently fail to handle. Rather than expanding dataset scale broadly, the curation strategy focuses on selecting samples that contain challenging but operationally important visual structures such as rotated field labels, logo-embedded vendor names, multi-column layouts, and fine print regulatory text. The objective is to increase the density of failure-inducing patterns within a compact dataset, enabling efficient adaptation of general-purpose VLMs to financial workflows while keeping annotation and compute costs low.

## 3.1. Identification of Failure Scenarios

To identify the visual patterns that should be represented during dataset construction, annotators manually audited a held-out collection of financial documents and analyzed the systemic failure modes exhibited by prevailing OCR and vision–language models. Annotators compared the model transcriptions against the ground truth and labeled the underlying cause of each error. This audit revealed a stable set of recurring breakdowns that accounted for the majority of transcription and layout inconsistencies, including vertical text orientation, fine print omissions, multi-column reading order errors, incomplete recognition of logo embedded fields, degraded scan quality, mixed handwritten and printed content, and the loss of hierarchical structural cues such as section headers or bullet lists. These failure scenarios form the basis for the subsequent construction of SAVIOR-Train and SAVIOR-Bench.

| Label | Percentage |
|---|---|
| Financial document | 50.566572% |
| Fine print | 40.014164% |
| Multi column | 23.796034% |
| Degraded image | 3.966006% |
| Handwritten text | 6.161473% |
| Hierarchical text | 1.203966% |
| Stylized logo | 0.354108% |
| Vertical text | 0.283286% |
| Horizontal text | 8.569405% |

Table 1. Attribute distribution in olmOCR-Bench. Critical failure modes (e.g., fine print, degraded images, stylized logos, vertical text, hierarchical text) make up only a small portion of the dataset.

## 3.2. Construction of SAVIOR-Train

Although olmOCR-Bench [24] is widely used for evaluating document OCR systems, its distribution of document attributes shows that critical failure modes are severely underrepresented. As shown in Table 1, the benchmark is dominated by standard financial documents and horizontal, multi-column layouts, whereas challenging cases that frequently cause VLM and OCR failures, such as fine print, degraded scans, stylized logos, vertical text, and hierarchical text, constitute only a small fraction of the dataset. This imbalance limits the benchmark's utility for assessing robustness in real enterprise environments.

To mitigate this, we introduce SAVIOR-Train, which contains 2,234 curated documents and OCR pairs selected to ensure broad coverage of the failure scenarios identified during the initial screening of financial documents. A large collection of documents was examined, and samples were added whenever their OCR predictions exhibited at least one of the recurring breakdown patterns described earlier. The selection process followed a balanced strategy in which documents were chosen to maintain representation across all failure scenarios rather than concentrating on any single pattern. Although the dataset is composed primarily of real documents, a small number of controlled synthetic variations were introduced specifically to increase coverage of extreme vertical text, which occurred infrequently in the source data but was consistently associated with severe transcription disruption. These additions affect only text orientation and do not introduce artificial content or change document semantics.

| Failure Mode | Count | Percentage |
|---|---|---|
| Vertical text | 200 | 9.0% |
| Fine print | 100 | 4.5% |
| Multi-column layouts | 150 | 6.7% |
| Logo-embedded text | 500 | 22.4% |
| Degraded image quality | 200 | 9.0% |
| Handwritten content | 50 | 2.2% |
| Structured headers | 1,000 | 44.8% |
| Bold / emphasized text | 1,034 | 46.3% |

Table 2. Distribution of failure modes represented in SAVIOR-Train. Categories are non-exclusive and a single document may exhibit multiple failure modes.

The distribution of these scenarios is summarized in Table 2. Categories are not exclusive because many documents contain more than one source of difficulty, for example, vendor names inside logos, along with small regulatory clauses or handwritten notes added to printed financial tables. This overlap is an intentional characteristic of SAVIOR-Train and reflects the nature of real financial documents, where challenging elements rarely appear in isolation. By concentrating multiple failure-inducing patterns inside a compact dataset and supplementing vertical orientation only when it was underrepresented, SAVIOR-Train provides a training signal that directly targets observed weaknesses without requiring large-scale corpus expansion.

### 3.3. SAVIOR-Bench

SAVIOR-Bench is a test dataset comprising 509 documents drawn from real-world business workflows. Each document is annotated to evaluate not only token-level transcription accuracy but also the preservation of semantic content and structural layout that are essential for downstream LLM-based processing. To capture the distribution of failure scenarios represented in SAVIOR-Train, SAVIOR-Bench incorporates material from five primary sources- UCSF(46.95%) [3], inhouse-synthetic (17.88%), miscellaneous real-world public documents (17.49%), Inv3D(9.04%) [7],and FUNSD(8.64%) [12]. enabling rigorous evaluation of a model's ability to handle enterprise-specific challenges such as layout fidelity, semantic segmentation, and regulatory content extraction. Evaluation on SAVIOR-Bench therefore emphasizes the end-to-end impact of OCR quality on downstream tasks, including structured field extraction, compliance analysis, and document classification.

To ensure that the benchmark provides reliable reference annotations for these tasks, we measured agreement across annotators who independently transcribed and structured each document. Inter-annotator agreement (IAA) was computed on the OCR transcriptions produced by two annotators per document using fuzzy string similarity based on the normalized Levenshtein ratio. Double annotations were collected for a subset of 187 documents because duplicating annotations across the full benchmark is impractical for long and complex financial documents, and we observed a mean similarity score of 0.761, indicating moderate to strong consistency in transcription quality. A histogram of similarity scores is shown in Figure 3, and summary statistics are provided in Table 3.

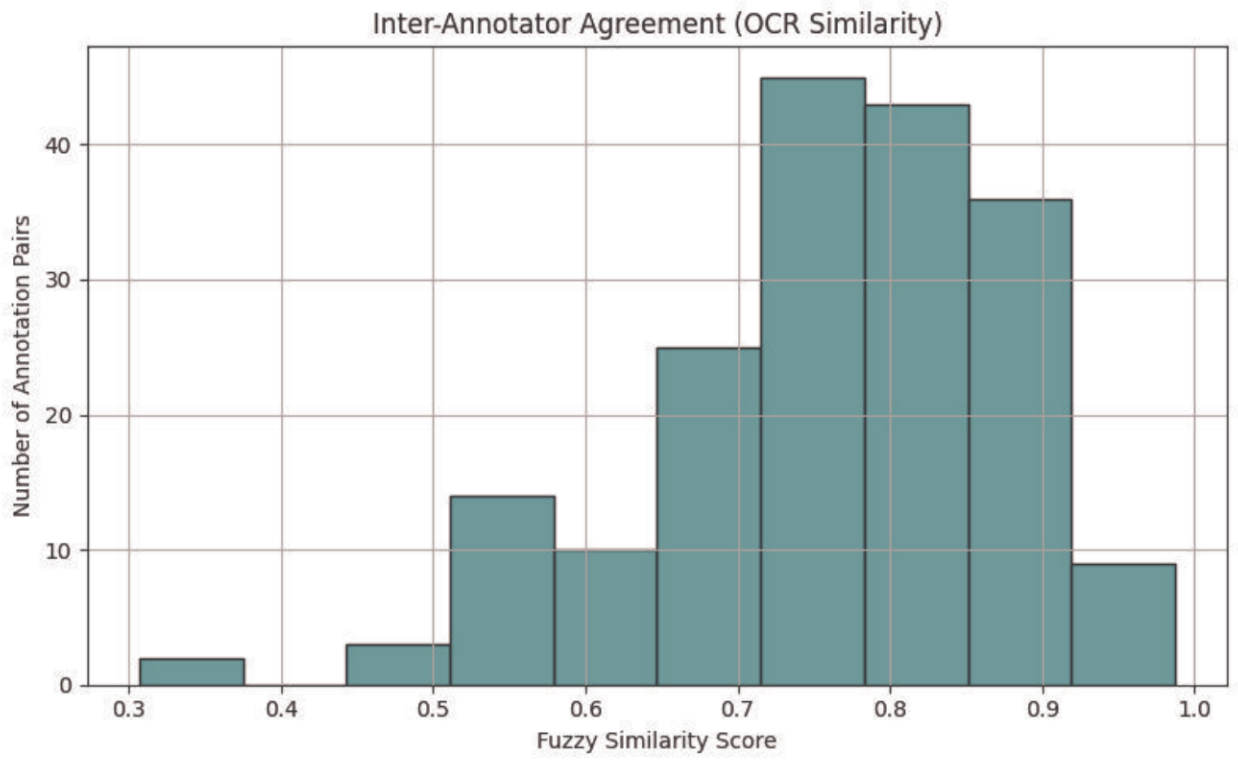

Figure 3. Histogram of inter-annotator similarity scores (OCR annotations).

| **Metric** | **Value** |
|---|---|
| Number of annotation pairs | 187 |
| Mean similarity | 0.761 |
| Minimum similarity | 0.307 |
| Maximum similarity | 0.987 |
| Standard deviation | 0.117 |

Table 3. Inter-Annotator Agreement (IAA) on OCR annotations using fuzzy string similarity.

### 3.4. Prompt Design for Training

Qwen2.5-VL-3B-Instruct and Qwen2.5-VL-7B-Instruct were fine-tuned using a structured prompt template composed of ten sequential instructions. The template explicitly prompts the model to *act as an OCR assistant* establishing consistent formatting expectations for all OCR outputs. The prompt guides the model to: (1) perform optical character recognition on the provided input image,

[3] UCSF Industry Documents Library. Accessed December 9, 2025. `https://www.industrydocuments.ucsf.edu/`

(2) preserve relevant formatting and layout cues from the source document, (3) generate clean textual output without meta-level commentary, (4) identify and tag logo text using XML-style delimiters, (5–6) apply Markdown conventions for bold text and header notation, (7) interpret and reproduce multi-column layouts, (8–9) process tables and other forms of structured data, and (10) manage edge cases that arise during text extraction.

## 4. Evaluation Metrics

We evaluate models on SAVIOR-Bench and report metrics designed to assess both semantic completeness and structural fidelity.

### 4.1. Word-Level Recall

Word-level recall quantifies the proportion of ground-truth words correctly recovered by the OCR system, counting each occurrence separately. Let $G$ denote the multiset of ground-truth words and $P$ the multiset of predicted words. For each $w \in G$, a match is counted if $w \in P$, and one instance of $w$ is then consumed from $P$. Recall is defined as:

$$\text{Recall} = \frac{1}{|G|} \sum_{w \in G} \mathbf{1}\{ w \in P \}, \tag{1}$$

where $\mathbf{1}\{\cdot\}$ is the indicator function.

We adopt word-level recall instead of character-level accuracy, since downstream applications rely on complete tokens. For example, predicting "Totl" instead of "Total" may be close at the character level, but it fails entirely in a key–value extraction task. Word-level recall thus better reflects semantic fidelity in enterprise document workflows.

### 4.2. Word-Level Precision

Word-level precision measures the proportion of predicted words that correctly correspond to ground truth, counting each occurrence separately. Let $P$ denote the multiset of predicted words and $G$ the multiset of ground-truth words. For each $w \in P$, a match is counted if $w \in G$, and one instance of $w$ is then consumed from $G$. Precision is defined as:

$$\text{Precision} = \frac{1}{|P|} \sum_{w \in P} \mathbf{1}\{ w \in G \}, \tag{2}$$

where $\mathbf{1}\{\cdot\}$ is the indicator function.

### 4.3. Layout Similarity Metric - PAIRS

While existing metrics capture transcription quality, they do not account for whether the spatial structure of the document is preserved. In enterprise workflows, layout fidelity is critical: an output that contains the correct words but misplaces them can render key–value extraction or tabular parsing unusable. For example, if an invoice contains "Total: $1,500" but the text is misplaced near the amount to be paid instead of the discount amount, word-level recall would still rate the transcription as correct. However, downstream LLMs for key–value extraction would misinterpret the field, resulting in incorrect mappings in enterprise workflows. We therefore introduce a layout similarity metric that quantifies structural accuracy. An example of this spatial encoding is shown in Fig. 4.

We represent both ground truth and predictions as 2D arrays. Each document is divided into lines, and within each line, words are placed into horizontal cells. Every word $i$ is assigned coordinates $(x_i, y_i)$, where $x_i$ denotes the index of the cell containing the center character of the word, and $y_i$ is the line number.

This representation makes it possible to compare word positions consistently between prediction and ground truth. For each unordered pair $(i, j)$, $1 \le i < j \le N$, we compute horizontal, vertical, and Euclidean distances in the ground truth,

$$\begin{aligned} h_{ij} &= |x_j - x_i|, \\ v_{ij} &= |y_j - y_i|, \\ d_{ij} &= \sqrt{h_{ij}^2 + v_{ij}^2}, \end{aligned} \tag{3}$$

and analogously $h'_{ij}$, $v'_{ij}$, $d'_{ij}$ in the prediction.

The relative error for each pair is

$$e_{ij} = \frac{|d'_{ij} - d_{ij}|}{d_{ij} + \epsilon}, \tag{4}$$

which we convert into a bounded similarity score

$$s_{ij} = 1 - e_{ij}^2, \tag{5}$$

where the quadratic penalty ensures that larger deviations are penalized more strongly. The raw layout similarity is then

$$S_{\text{raw}} = \frac{2}{N(N-1)} \sum_{1 \le i < j \le N} s_{ij}. \tag{6}$$

To normalize across documents with different spatial scales, we compute $z$-scored pairwise distances

$$z_{ij} = \frac{d_{ij} - \mu}{\sigma}, \qquad z'_{ij} = \frac{d'_{ij} - \mu}{\sigma}, \tag{7}$$

where $\mu$ and $\sigma$ denote the mean and standard deviation of the set $\{d_{ij}\}$ computed from the ground truth.

The normalized layout similarity is then defined as

$$S_z = \frac{2}{N(N-1)} \sum_{1 \le i < j \le N} \left( 1 - \left( \frac{|z'_{ij} - z_{ij}|}{|z_{ij}| + \epsilon} \right)^2 \right), \tag{8}$$

This metric directly evaluates whether relative word positions are preserved, complementing word-level recall by capturing structural fidelity in OCR outputs.

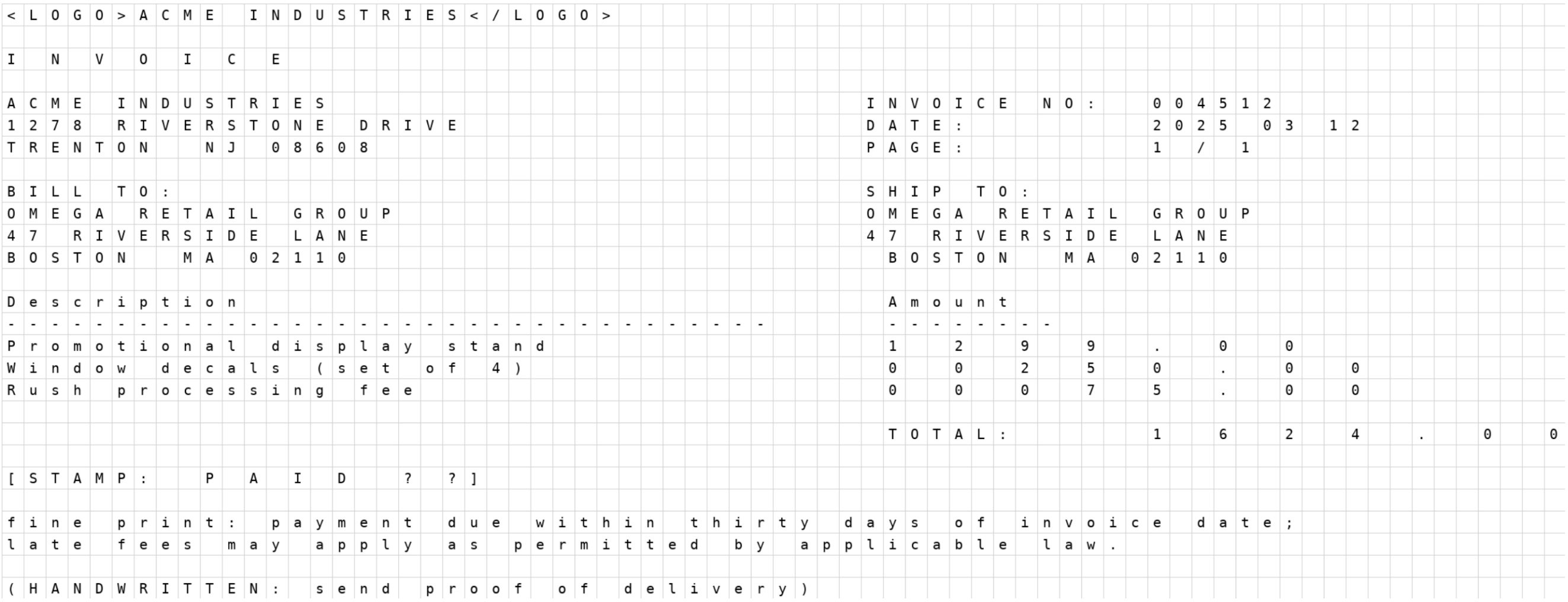

Figure 4. Visualization of the 2D spatial grid representation used in PaIRS. Each word is positioned according to its row and horizontal cell index, enabling structural comparison between predictions and ground truth beyond transcription accuracy.

# 5. Training Strategy

Our fine-tuning methodology is designed for sample-efficient adaptation by strategically targeting key model components. We apply Low-Rank Adaptation (LoRA) with a rank of 32 and scaling factor $\alpha = 64$ to two specific areas: the final layer of the vision encoder and the language model decoder. This targeted approach for fine-tuning the last encoder layer allows the model to refine its high-level visual feature extraction for enterprise-specific challenges, such as stylized logos and fine print. Simultaneously, strengthening the decoder is crucial for improving its robustness in generating accurate text from degraded or noisy inputs, such as poor-quality scans.

## 5.1. Implementation Details

The vision encoder and language model are optimized with learning rates of $1 \times 10^{-4}$ and $5 \times 10^{-4}$ respectively. Training uses a batch size of 1 with gradient accumulation over 10 steps. The 3B and 7B variants of Qwen2.5-VL-Instruct were fine-tuned for 7 and 5 epochs respectively. All models are trained on 2 NVIDIA H100 80GB GPUs using CUDA 12.3. We evaluate models on SAVIOR-Bench, a 509-document benchmark annotated by domain experts. Inference is performed with top-p $=$ 0.9 and temperature $=$ 0.1.

## 5.2. Baseline Models

We include commercial and open-source OCR systems such as Nanonets-OCR-s and PaddleOCR 3.0 as baselines. These systems were evaluated in their native inference settings without fine-tuning on SAVIOR-Train to reflect realistic deployment conditions rather than adapted usage.

Unlike Qwen-based models, which generate a linear sequence of tokens in reading order, PaddleOCR returns a dictionary keyed by bounding boxes. These formats do not provide a strictly ordered word sequence, and they do not expose a controllable decoding interface that ensures one-token-per-cell generation. As a result, their outputs cannot be converted into the consistent 2D token grid required for PaIRS without building substantial custom adapters that would introduce implementation-specific bias. For this reason, layout similarity metrics such as PaIRS are reported only for models that can be prompted to produce sequential text. Nevertheless, markup-based systems are still evaluated to assess how closely their structured outputs resemble the visual layout of documents, providing partial insight into layout fidelity even when full PaIRS compatibility is not possible.

# 6. Results and Analysis

We evaluate different OCR models along four axes: transcription accuracy, layout preservation, downstream QA performance, and deployment efficiency under quantization. Corresponding metrics include word-level recall and precision, PaIRS layout similarity, F1 on document QA, and throughput of 4-bit models in constrained environments. Throughout this section, models with the “F” suffix denote variants fine-tuned using the SAVIOR-Train, specifically, Qwen2.5-VL-3B-F ( fine-tuned from Qwen2.5-VL-3B-Instruct), and Qwen2.5-VL-7B-F ( fine-tuned from Qwen2.5-VL-7B-Instruct).

## 6.1. Recall and Precision

As shown in Table 4, Qwen2.5-VL-7B-F, fine-tuned with SAVIOR, achieves the highest overall performance on SAVIOR-Bench, reaching a word-level recall of 0.9294 and a precision of 0.9157. This substantially exceeds both open-source and commercial baselines, including Nanonets-OCR-s (0.9039 recall / 0.5336 precision) and PaddleOCR 3.0 (0.8684 recall / 0.8924 precision). Despite being trained on only 2,234 curated samples, the model displays strong generalization, underscoring the sample-efficiency of the SAVIOR methodology.

| Model | Recall | PaIRS | Precision |
|---|---|---|---|
| **Prompt-Controlled** | | | |
| GPT-4o | 0.8854 | 0.5826 | 0.9322 |
| Gemini-2.5-Flash | 0.9144 | 0.6074 | 0.8967 |
| Qwen2.5-VL-7B-Instr. | 0.9091 | 0.5629 | 0.9145 |
| Qwen2.5-VL-3B-Instr. | 0.8547 | 0.5764 | 0.9120 |
| **Non-Prompt-Controlled** | | | |
| DeepSeek-OCR | 0.7706 | 0.5034 | 0.9008 |
| Mistral OCR | 0.8611 | 0.5382 | 0.8986 |
| dots.ocr | 0.8992 | 0.5540 | 0.9290 |
| olmOCR-7B-0725-FP8 | 0.7463 | 0.5864 | 0.9450 |
| PaddleOCR 3.0 | 0.8684 | N/A | 0.8924 |
| Nanonets-OCR-s | 0.9039 | 0.4354 | 0.5336 |
| PaddleOCR-VL | 0.5955 | 0.4180 | 0.8194 |
| Qwen2.5-VL-7B-F | **0.9294** | 0.7066 | **0.9157** |
| Qwen2.5-VL-3B-F | 0.9167 | **0.7068** | 0.9101 |

Table 4. Performance comparison on SAVIOR-Bench. The PaIRS metric evaluates structure-aware fidelity. Prompt-Controlled models can be prompted to produce structured output compatible with the PaIRS 2D grid, whereas Non–Prompt-Controlled models cannot be prompted due to their fine-tuned, fixed output templates.

The smaller Qwen2.5-VL-3B-F variant achieves a competitive recall of 0.9167 with a precision of 0.9101, while larger or instruction-tuned variants yield only marginal gains at substantially higher computational cost. These results indicate that the 7B fine-tuned model provides the most favorable balance between accuracy and efficiency for practical deployment in enterprise OCR workflows.

## 6.2. PaIRS metric for layout preservation

Table 5 reports raw layout similarity scores across Euclidean, horizontal, and vertical distances. The fine-tuned Qwen2.5-VL models achieve the strongest structural alignment overall: Qwen2.5-VL-3B-F attains the highest Euclidean similarity (0.8214) and vertical similarity (0.8945), while Qwen2.5-VL-7B-F achieves the best horizontal similarity (0.6978). Gemini-2.5-Flash and GPT-4o form a middle tier, outperforming smaller open-source OCR systems but trailing both fine-tuned Qwen variants. Mistral-OCR exhibits the weakest structural preservation, particularly in the horizontal dimension.

Table 6 reports layout similarity after z-score normalization to account for scale variation across documents. The relative rankings remain consistent: Qwen2.5-VL-3B-F and Qwen2.5-VL-7B-F lead across all dimensions, with the 3B model slightly ahead in Euclidean and vertical similarity and the 7B model leading in horizontal similarity. Vertical alignment remains more stable than horizontal alignment even after normalization, indicating that most OCR systems preserve line ordering more reliably than within-line spacing.

Finally, it is important to interpret cross-model comparisons with caution: models producing different output formats (e.g., JSON, Markdown, or free-form text) yield varying spatial token arrangements when projected into 2D

| Model | Euclidean | Horizontal | Vertical |
|---|---|---|---|
| DeepSeek-OCR | 0.6121 | 0.4537 | 0.6965 |
| Mistral-OCR | 0.6526 | 0.5175 | 0.7456 |
| dots.ocr | 0.6912 | 0.5451 | 0.7325 |
| olmOCR-7B-0725-FP8 | 0.7006 | 0.5756 | 0.7831 |
| GPT-4o | 0.7140 | 0.5483 | 0.7847 |
| Gemini-2.5-Flash | 0.7560 | 0.5803 | 0.7781 |
| Nanonets-OCR-s | 0.5277 | 0.5117 | 0.4506 |
| PaddleOCR-VL | 0.4965 | 0.4203 | 0.4965 |
| Qwen2.5-VL-7B-Instr. | 0.7068 | 0.5403 | 0.7489 |
| Qwen2.5-VL-3B-Instr. | 0.7093 | 0.5541 | 0.7965 |
| Qwen2.5-VL-7B-F | 0.8200 | **0.6978** | 0.8895 |
| Qwen2.5-VL-3B-F | **0.8214** | 0.6940 | **0.8945** |

Table 5. Raw PaIRS layout similarity scores across Euclidean, horizontal, and vertical dimensions (a higher score is better). Fine-tuned Qwen2.5-VL models demonstrate the strongest structural alignment, with Qwen2.5-VL-3B-F and Qwen2.5-VL-7B-F outperforming all baselines across all metrics.

| Model | Euclidean | Horizontal | Vertical |
|---|---|---|---|
| DeepSeek-OCR | 0.3387 | 0.3736 | 0.5457 |
| Mistral-OCR | 0.3704 | 0.3732 | 0.5702 |
| dots.ocr | 0.3923 | 0.4065 | 0.5566 |
| olmOCR-7B-0725-FP8 | 0.4198 | 0.4340 | 0.6057 |
| GPT-4o | 0.4164 | 0.4148 | 0.6178 |
| Gemini-2.5-Flash | 0.4567 | 0.4506 | 0.6231 |
| Nanonets-OCR-s | 0.3296 | 0.3965 | 0.3963 |
| PaddleOCR-VL | 0.3151 | 0.3423 | 0.4371 |
| Qwen2.5-VL-7B-Instr. | 0.4032 | 0.4162 | 0.5622 |
| Qwen2.5-VL-3B-Instr. | 0.3993 | 0.4176 | 0.5819 |
| Qwen2.5-VL-7B-F | **0.5557** | **0.5648** | 0.7119 |
| Qwen2.5-VL-3B-F | 0.5548 | 0.5622 | **0.7140** |

Table 6. PaIRS layout similarity scores after $z$-score normalization to account for scale differences across documents (a higher score is better). Normalized rankings mirror the raw scores, with Qwen2.5-VL-F models maintaining the highest layout fidelity across all dimensions.

space, which can influence PaIRS scores independent of true recognition quality.

### 6.3. Performance on downstream QA task

Table 7 reports F1 scores on a downstream QA benchmark composed of 105 question–answer pairs across 50 documents sampled from SAVIOR-Bench, with questions specifically targeted toward known OCR failure modes. All predictions were generated using the Meta-Llama-3.3-70B-Instruct-AWQ-INT4[4] model, which was prompted to provide short, phrase-based answers using only the OCR output as context. Qwen2.5-VL-F models exhibit consistently strong transfer performance, with Qwen2.5-VL-7B-F achieving the highest F1.

| Model | F1 |
|---|---|
| DeepSeek-OCR | 0.6527 |
| Mistral-OCR | 0.7597 |
| dots.ocr | 0.8393 |
| olmOCR-7B-0725-FP8 | 0.6906 |
| GPT-4o | 0.7868 |
| Gemini-2.5-Flash | 0.8582 |
| Nanonets-OCR-s | 0.8663 |
| PaddleOCR-VL | 0.7409 |
| Qwen2.5-VL-7B-Instr. | 0.8393 |
| Qwen2.5-VL-3B-Instr. | 0.7758 |
| Qwen2.5-VL-7B-F | **0.8963** |
| Qwen2.5-VL-3B-F | 0.8418 |

Table 7. F1 scores on downstream QA task

### 6.4. Performance of 4-bit quantized models

| Model | Recall | Precision | PaIRS |
|---|---|---|---|
| Qwen2.5-VL-7B-F | **0.9170** | **0.9170** | **0.7073** |
| Qwen2.5-VL-3B-F | 0.8864 | 0.8827 | 0.6843 |

Table 8. Performance of 4-bit quantized Qwen2.5-VL-3B-F and Qwen2.5-VL-7B-F models.

To evaluate performance under memory constraints, we analyze 4-bit quantized variants of Qwen2.5-VL-3B-F and Qwen2.5-VL-7B-F models using merged-weight representations. As reported in Table 8, quantization leads to modest performance reductions. The 7B model shows a decrease in recall of 0.012 with minimal effect on layout similarity, while the 3B model exhibits a recall drop of 0.0303 but remains competitive given its substantially reduced memory footprint.

[4]https://huggingface.co/ibnzterrell/Meta-Llama-3.3-70B-Instruct-AWQ-INT4

### 6.5. Deployment of quantized models

To assess the deployment viability of 4-bit quantized Qwen2.5-VL-3B-F and Qwen2.5-VL-7B-F models in latency-sensitive environments, we measured token-level throughput using vLLM [15]. Across both 3B and 7B variants, quantization preserves high throughput (Table 9), with Qwen2.5-VL-3B-F model reaching up to 1085.81 tokens/s on an A100 80GB GPU.

| Model | GPU | VRAM (GB) | Throughput (tokens/sec) |
|---|---|---|---|
| Qwen2.5-VL-7B-F | A100 | 75.6 | 977.60 |
| Qwen2.5-VL-3B-F | A100 | 74.3 | **1085.81** |

Table 9. Memory (VRAM) and throughput (tokens/sec) for 4-bit quantized Qwen2.5-VL-3B-F and Qwen2.5-VL-7B-F models.

## 7. Conclusion and Future Work

SAVIOR is a sample-efficient data curation methodology designed to align Vision–Language Models (VLMs) with OCR tasks in enterprise environments. It targets common VLM failure modes such as vertical text, embedded logos, fine print, and degraded scans, demonstrating that focused high-utility curation can outperform scale-driven training.

We introduce SAVIOR-Bench, a 509-document benchmark with expert OCR annotations from the finance and accounting domain. We further propose PaIRS, a structure-aware evaluation metric that measures pairwise relational similarity between predicted and ground-truth word locations.

Our fine-tuned model surpasses PaddleOCR, PaddleOCR-VL, Mistral-OCR and Nanonets-OCR-s on SAVIOR-Bench while requiring substantially fewer training samples. SAVIOR-trained models also retain accuracy under 4-bit quantization, enabling deployment in low-memory, latency-sensitive settings.

Future work includes expanding SAVIOR to additional document types such as contracts and receipts, automating failure-mode discovery via model introspection, and combining SAVIOR-style curation with multimodal instruction tuning for downstream tasks such as key–value extraction and summarization.

**Acknowledgments**

This research was conducted at Hyperbots Inc. and annotations were carried out by commerce graduates under expert supervision. We thank the annotators: Aishwarya S. Warad, Ashwin Devkate, Mokshaa M. Omkar, Pavithra Prakash, Pratham Shyam, R. Chaitanya, Rohith M. Dhondaley, Surya Shobith, and Trisha L. R. for their contributions. Annotators were fairly compensated, and the study adhered to established ethical standards.